\documentclass[letterpaper, 10 pt, conference]{ieeeconf}
\IEEEoverridecommandlockouts                              

\usepackage{ifpdf}

\usepackage{xcolor}
\usepackage{color}

\usepackage{textcomp}

\usepackage[font=small]{subcaption} 

   \usepackage[pdftex]{graphicx}
   \graphicspath{{figures/}{../jpeg/}}
  \DeclareGraphicsExtensions{.pdf,.jpeg,.png}
\usepackage{amsmath}
\usepackage{esvect}
\usepackage{mathtools}
\usepackage{amssymb}
\usepackage{amsfonts}
\usepackage{mathabx}

\usepackage{array}
\usepackage[algo2e]{algorithm2e} 
\usepackage{algorithm}
\usepackage{algorithmic}

\usepackage{booktabs}

\usepackage{cite}

\usepackage{textcomp}

\begin{document}
\captionsetup[figure]{labelfont={bf},labelformat={default},labelsep=period,name={Fig.}}

\title{\LARGE \bf
Towards Safe Landing of Falling Quadruped Robots Using a 3-DoF Morphable Inertial Tail
}

\author{$\text{Yunxi Tang}^{\dagger}$, $\text{Jiajun An}^{\dagger}$, $\text{Xiangyu Chu}^{*}$, $\text{Shengzhi Wang}^{}$, $\text{Ching Yan Wong}^{}$, and $\text{K. W. Samuel Au}^{}$ 
\thanks{This work was supported in part by the Chow Yuk Ho Technology Centre of Innovative Medicine, The Chinese University of Hong Kong; in part by the Multiscale Medical Robotics Centre, AIR@InnoHK; and in part by the Research Grants Council (RGC) of Hong Kong under Grant 14209719. All authour are with Department of Mechanical and Automation Engineering, The Chinese University of Hong Kong, Hong Kong, China. $\dagger$: Equal Contribution. $*$: Corresponding author.
}
}

\maketitle
\thispagestyle{empty}
\pagestyle{empty}

\begin{abstract}
Falling cat problem is well-known where cats show their super aerial reorientation capability and can land safely. For their robotic counterparts, a similar falling quadruped robot problem, has not been fully addressed, although achieving safe landing as the cats has been increasingly investigated. Unlike imposing the burden on landing control, we approach to safe landing of falling quadruped robots by effective flight phase control. Different from existing work like swinging legs and attaching reaction wheels or simple tails, we propose to deploy a 3-DoF morphable inertial tail on a medium-size quadruped robot. In the flight phase, the tail with its maximum length can self-right the body orientation in 3D effectively; before touch-down, the tail length can be retracted to about 1/4 of its maximum for impressing the tail's side-effect on landing. To enable aerial reorientation for safe landing in the quadruped robots, we design a control architecture, which has been verified in a high-fidelity physics simulation environment with different initial conditions. Experimental results on a customized flight-phase test platform with comparable inertial properties are provided and show the tail's effectiveness on 3D body reorientation and its fast retractability before touch-down. An initial falling quadruped robot experiment is shown, where the robot Unitree A1 with the 3-DoF tail can land safely subject to non-negligible initial body angles.

\end{abstract}

\section{Introduction}
A sequence of motions such as aerial reorientation and stable landing have been investigated in both animals \cite{KaneFallingCatPhenomenon1969,FukushimaSquirrels2021} and robots \cite{KurtzMiniCheetah2022, RudinCatLike2022}. Such motions are critical for safety and survival when animals or robots are subject to unexpected falls. For example, a falling cat can rotate its front and back bodies, swing its tail and legs to self-right before safe landing with four feet pointing downwards \cite{KaneFallingCatPhenomenon1969}. Squirrels that were catapulted off a track could stabilize themselves using tail motion, allowing them to land successfully \cite{FukushimaSquirrels2021}. In robots, especially for medium-size quadruped robots like Mini Cheetah and Unitree A1, they may suffer from the same safety issues in falling. Referring to the famous falling cat problem, we can call it \textit{falling quadruped robot problem}. Thus, landing the quadruped robots safely needs to be solved. 

\begin{figure}
    \centering
    \includegraphics[width=50mm]{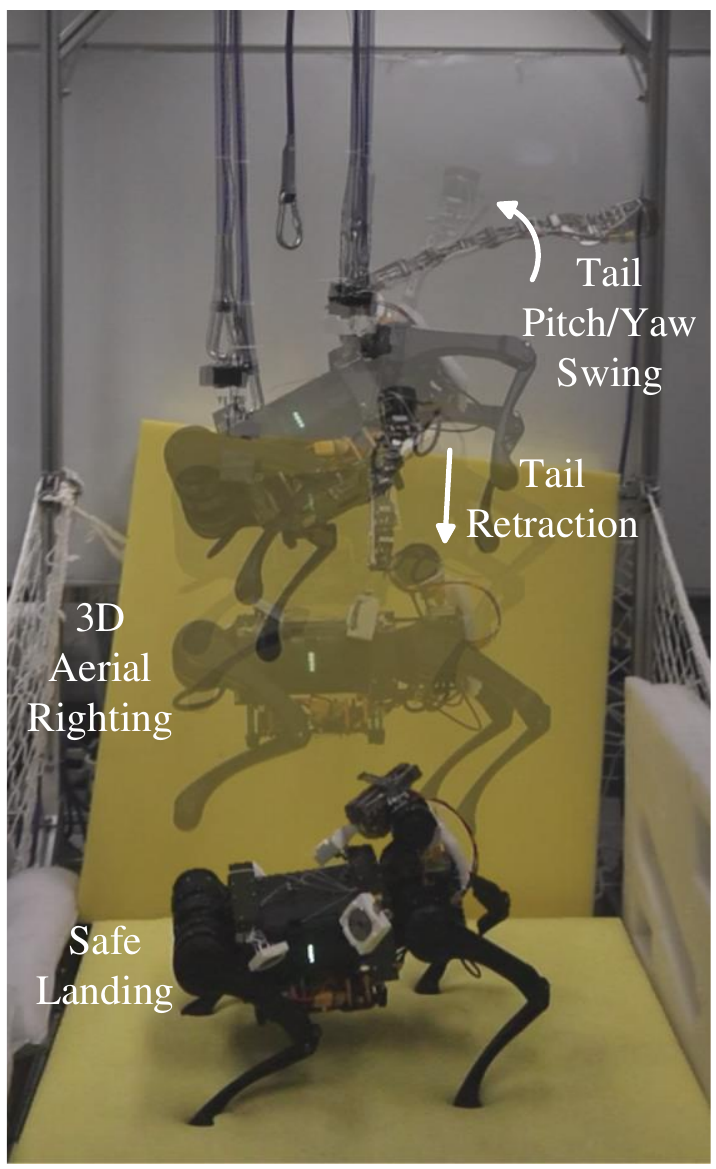}
    \caption{A combined motion snapshot of Unitree A1 performing 3D aerial righting and safe landing by utilizing a 3-DoF morphable inertial tail in a fall from 1 m height.}
    \label{fig:system_integration}
\end{figure}

There are two paradigms to approach such a problem: 1) designing landing strategies to search optimal contact sequence and optimize contact forces for landing impact; 2) using limbs or extra appendages to right the body to a horizontal pose and then applying a simple landing controller. In the state-of-the-art work \cite{SHJ}, the first paradigm has been implemented, but their results show that Mini Cheetah can only handle \textit{horizontal} drops in hardware. Besides, the robot may be damaged because of uneven leg force distribution at touch-down when the body has significant orientation offset from the horizontal. The second paradigm aims to reorient the body to the horizontal even the desired pose (accommodating the terrain and environment) before touching down. This alleviates the burden of landing control and mitigates robots' mechanical damage. In this paper, we will focus on the second paradigm and integrate a 3-DoF tail into a quadruped robot to enhance the capability of safe landing. 

Although, recently, a few efforts have been made to acquire the reorientation-for-land capability in quadruped robots, their performance is still far from that of their biological counterparts. The work \cite{KurtzMiniCheetah2022} has enabled, a big robotic cat, Mini-Cheetah to land on its feet from falls with initial pitch within $\pm 90^{\degree}$, but the motion was constrained in sagittal plane. Similarly, \cite{RudinCatLike2022} presented a combination of 2D reorientation and landing locomotion behaviors based on a physical quadruped robot SpaceBok, although the same behaviors in 3D were implemented in simulation. Within the mentioned work, the leg swinging is not effective in inducing angular momentum change, because 1) the Moment of Inertia (MoI) of the legs is relatively small, compared with that of bodies; 2) the workspace of the legs is limited and it may consume more time, compared with that of tails which are common in many quadruped animals. These limitations also explain why \cite{KurtzMiniCheetah2022} added additional mass to the foot for increasing the leg's MoI and \cite{RudinCatLike2022} assumed the drop happened in a low gravity environment for increasing aerial duration.

Except for using legs \cite{KurtzMiniCheetah2022,RudinCatLike2022, GosselinReorientation2022}, reaction wheels and tails have been included in quadruped robots for enhancing locomotion capability in both flight and stance phases. \cite{KolvenbachMoon2019, ZacharyCMU2022, Roscia2022} used reaction wheels to assist locomotion, however, \cite{KolvenbachMoon2019} can only stabilize the pitch direction in the flight phase, and \cite{ZacharyCMU2022, Roscia2022} showed the aerial reorientation capability in simulations although they built prototypes. In terms of the tails, a simple application is using a tail to reject disturbance along pitch \cite{YangCMU2021}, yaw \cite{FawcettArticulatedTails2021}, roll \cite{BriggsTails2012} directions. \cite{YangCMU2022} used a 2-DoF tail with pitch and yaw control capabilities to react to elevation changes; more specifically, the tail's cone motion protected the robot from tipping when falling off a cliff. Besides, some researchers used a tail for airborne righting and successful land. \cite{NorbyAerodynamic2021} designed an inertial tail with aerodynamic drag to allow a quadruped robot Minitaur to reorient from a $90$ degree pitch angle before landing. \cite{LiuSerpentine2021} proposed to use a serpentine robotic tail to stabilize body's pitch and roll to zero while landing. Among the work related to the tailed quadruped robots, only a few of them have provided hardware verification, especially focusing on planar (aerial) motion \cite{BriggsTails2012, NorbyAerodynamic2021}. Although \cite{LiuICRA2022} only built a reduced complexity quadruped robot designed for studying the serpentine robotic tail, no experimental results on the tailed robot have been provided. Considering multiple functions of the tail in quadruped robots, here we will constrain our attention on the reorientation-for-land capability and the study of improving forward velocity (e.g., in \cite{Heim2016}) or facilitating sharp turning (e.g., in \cite{Patelturning2013}) will be our future interest.

In this paper, we propose to integrate a 3-DoF morphable inertial tail (pitch, yaw, and telescoping) into a quadruped robot for enabling 3D aerial reorientation and then inducing safe landing. To our best knowledge, only a few 3-DoF tails have been designed \cite{AnMorphable2020, AnMorphable2022}, one of which \cite{AnMorphable2022} investigated the use of the 3-DoF tail for somersault motion with a twist, but only a small-size tethered monopod robotic platform was used.
Although a 2-DoF tail is commonly used for 3D aerial reorientation, e.g., \cite{ChuNSA2019}, there is a conflict between \textit{aerial reorientation} and \textit{landing balance} using a 2-DoF tail, because the 2-DoF tail configuration (or location) at the end of reorientation is uncontrolled and varies as different initial body angles. However, the tail configuration during stance phase has preference such that the tail's collision with the ground should be avoided and minimal disturbance would be imposed on body balance. 
To this end, for the first time, we introduce the 3-DoF tail to a quadrupedal robot where the tail with the maximal length (degenerating to a 2-DoF tail) can be used for self-righting in 3D effectively. Also, the tail can be retracted before touch-down for impressing the tail's side-effect and increasing the landing success. What we want to emphasize is the 3-DoF tail is designed to be modular and potentially available for other robots.

The contributions of this paper are: 
1) We integrate a 3-DoF morphable inertia tail into a quadrupedal robot and the tail can increase the quadrupedal robot's 3D aerial righting capability for safe landing. In experiments, the tail can help the robot adjust from a large 3D inclined posture to a desired posture during falling, which provides good preparation for the following safe landing task.
2) To reduce the potential damage to the quadrupedal robot, we design a flight-phase test platform that has a similar size and weight to the quadrupedal robot (Unitree A1) for initial experiments. 
Experimental results on the platform show the tail's effectiveness on 3D body reorientation and its fast retraction speed ($\sim 2$ m/s) before touch-down.
3) We complete a consecutive large 3D reorientation (zeroing $30^{\degree}$ pitch and $30^{\degree}$ roll offsets, and keeping yaw zero) and safe landing motion on the tailed A1 robot from $1$ m height.

\section{System Modelling}
\begin{figure}[t]
    \centering
    \includegraphics[width=60mm]{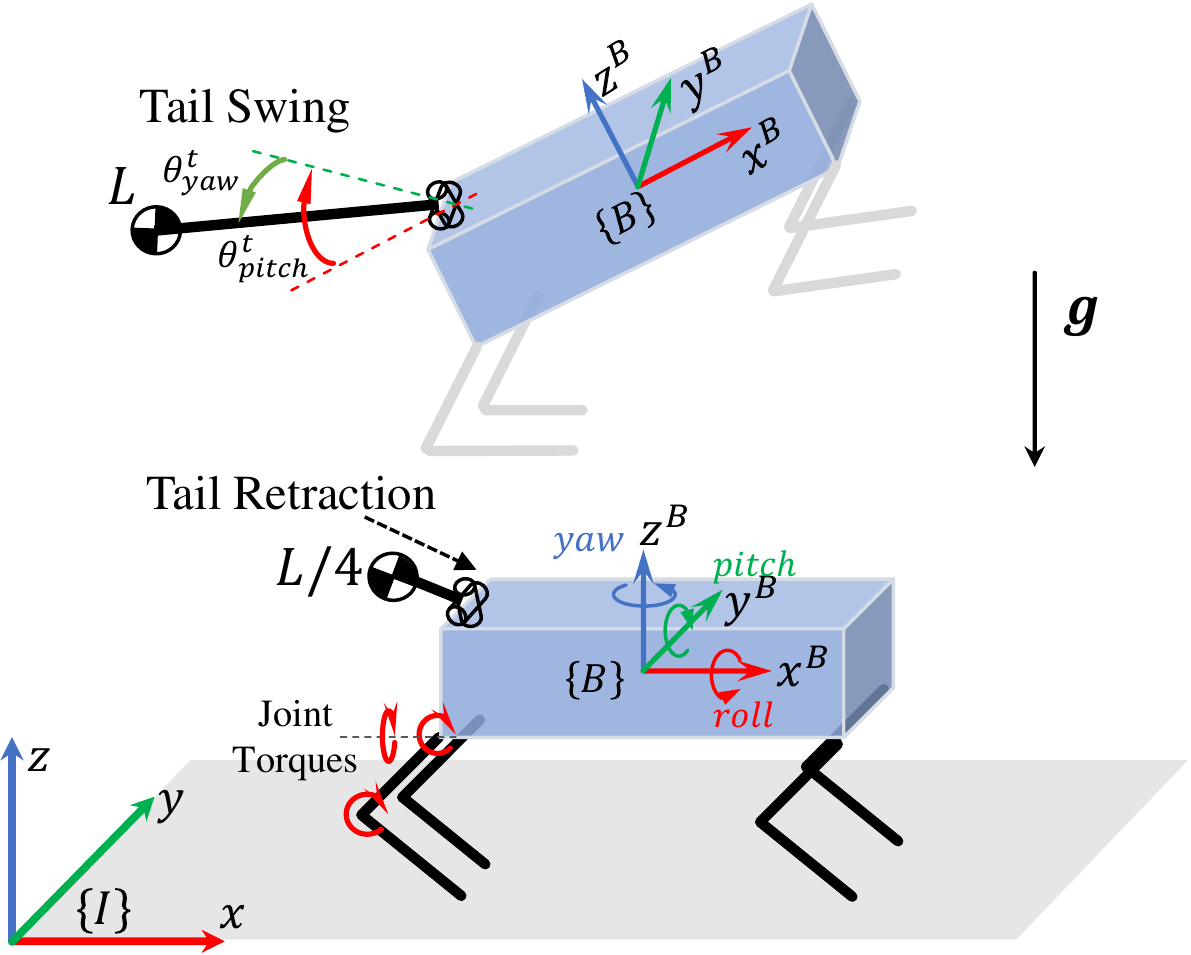}
    \caption{Simplified system model of the tailed quadrupedal robot.}
    \label{fig:tSRBD}
\end{figure}

\begin{figure*}[t]
    \centering
    \includegraphics[width=130mm]{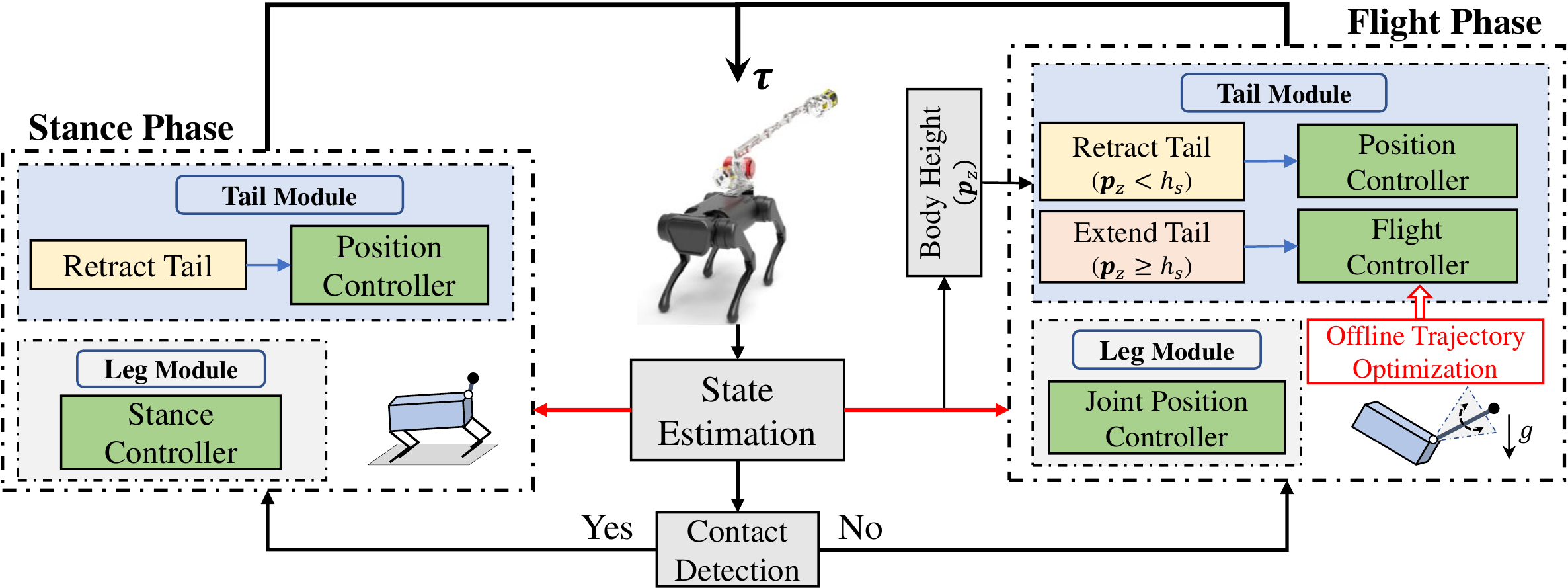}
    \caption{Planning and control framework for the tailed quadruped robot. An offline trajectory optimization is employed for the aerial reorientation and an additional flight tracking controller is designed as shown in the \textbf{Flight Phase} block. $h_s$ is the remained height after reoreintation. A PD stance controller is used to keep the robot balance as shown in the \textbf{Stance Phase} block.}
    \label{Control_Overview}
\end{figure*}
The motion patterns of a tailed quadrupedal robot in the flight phase and stance phase are different. As mentioned before, the tail will keep its maximum length for effective body reorientation in most of the flight phase. The 3-DoF morphable tail actually degenerates to a 2-DoF tail in airborne and thus we can simplify the system as a low-dimensional model, \textit{tailed Single Rigid Body Dynamics} model (tSRBD) as shown in Fig.~\ref{fig:tSRBD}. 
$L$ is the maximum tail length. 
$\theta^t_{pitch}$ and $\theta^t_{yaw}$ are the tail swing angles along the pitch and yaw directions, respectively.
We only focus on the tail's usage and assume the leg joints are kept at a proper configuration for landing since the legs' small weight and MoI are ineffective in aerial righting. 
Referring to the conventions in \cite{RDLN}, the system state of tSRBD is defined as,
\begin{equation}
\begin{aligned}
    \boldsymbol{q}_f &:= \begin{bmatrix}
    \boldsymbol{p} & \boldsymbol{\Theta} &\boldsymbol{q}_t
    \end{bmatrix} \in SE(3) \times \mathbb{R}^2,\\
    \quad \boldsymbol{u}_f &:=\begin{bmatrix}
    \boldsymbol{\dot{p}} &\boldsymbol{\omega} &\boldsymbol{\dot{q}}_t
    \end{bmatrix}\in \mathbb{R}^8,
\end{aligned}
\end{equation}
where $\boldsymbol{p}=[\boldsymbol{p}_x,\boldsymbol{p}_y,\boldsymbol{p}_z]$ is the position of the body's center of mass (CoM) and $\boldsymbol{q}_t=[\theta^t_{pitch},\theta^t_{yaw}]$ is the tail's joint positions. $\boldsymbol{\Theta}=[\boldsymbol{q}_x,\boldsymbol{q}_y,\boldsymbol{q}_z,\boldsymbol{q}_w]$ is the unit quaternion representation of the body orientation. 
Note that the body's position, orientation, and linear velocity are represented in the inertial frame $\{I\}$.
The body's angular velocity $\boldsymbol{\omega}$ is expressed in the base coordinates $\{B\}$. 
The equations of motion (EoM) can be written as,
\begin{equation}{}
\boldsymbol{M}_f(\boldsymbol{q}_f)\boldsymbol{\dot{u}}_f+\boldsymbol{b}_f(\boldsymbol{q}_f,\boldsymbol{u}_f) +\boldsymbol{g}_f(\boldsymbol{q}_f)=\boldsymbol{S}^T\boldsymbol{\tau},
\end{equation}
where $\boldsymbol{M}_f$ is the inertia matrix, $\boldsymbol{b}_f$ is the Coriolis and centrifugal terms, and $\boldsymbol{g}_f$ is the gravitational term. $\boldsymbol{\tau}$ is the joint torques of the tail. $\boldsymbol{S}$ is the selection matrix representing the under-actuation of the base,
$$
    \boldsymbol{S}=\begin{bmatrix}
    \boldsymbol{0}_{2\times 6} &
    \boldsymbol{I}_{2 \times 2}
    \end{bmatrix} \in \mathbb{R}^{2 \times 8}.
$$
After an effective aerial reorientation, the tail will quickly retract to $1/4$ of its maximum length (Fig.~\ref{fig:tSRBD}) before landing and the robot has similar mass distribution to its original state.
In this paper, we focus on showing the paradigm of reducing landing control burden via effective flight phase control and thus a controller specific to landing (e.g., \cite{SHJ}) will not resort. 
Therefore, a stance-phase dynamic model is not specified as a simple PD leg controller works for the stance phase safe landing task.

\section{Control Framework}

As a single controller is challenging in controlling a hybrid system, we develop a control framework to achieve the \textit{falling quadruped robot} task in this section. The planning and control framework is shown in Fig.~\ref{Control_Overview}. The whole motion is divided into two phases: flight phase and stance phase. In the reorientation phase, the robot adjusts its body orientation via swinging the tail with its maximum length. Then, the tail will be retracted close to the body for landing preparation after body self-righting. When contact is detected, the robot mainly uses its legs instead of the tail to keep balance on the ground in the stance phase. Each phase can use different controllers in corresponding blocks (green blocks in Fig.~\ref{Control_Overview}). In this paper, we select a trajectory optimization based controller for the flight phase and a compliant joint PD controller for the stance phase.

\subsection{Trajectory Optimization for Aerial Reorientation}
To realize the reorientation task of the tailed quadruped robot, the internal dynamics (conversation of angular momentum) can be utilized to adjust the body orientation in airborne. Trajectory optimization (TO) is an effective way to plan trajectories or design controllers by exploiting system dynamics and incorporating state/control constraints. Here, we adopt the TO method to obtain an optimized trajectory offline given the height and initial configuration of the robot, and the optimal trajectories provides a safe reorientation reference due to the satisfaction of physical constraints. Specifically, a custom-made differential dynamic programming (DDP) solver is employed in the offline stage. More details of the solver can be found in \cite{HM-DDP}. The trajectory optimization problem is formed as 
\subsubsection{Objective Function}
    \begin{equation}
        J(\boldsymbol{x}_{0},\boldsymbol{\tau}_{0:N-1}) = \sum_{k=0}^{N-1} \ell (\boldsymbol{x}_k,\boldsymbol{\tau}_{k}) + \ell _{f}(\boldsymbol{x}_N),
    \end{equation}
    where $N$ is the horizon length and $\boldsymbol{x}_0$ is the given initial state. State $\boldsymbol{x}_k$ at each time step is,
    $$
    \boldsymbol{x}_k=\begin{bmatrix}
    \boldsymbol{p} &\boldsymbol{\Theta} &\boldsymbol{q}_t &
    \boldsymbol{\dot{p}} &\boldsymbol{\omega} &\boldsymbol{\dot{q}}_t
    \end{bmatrix}_k,
    $$
    and the running and terminal objective objective functions, $\ell (\boldsymbol{x}, \boldsymbol{\tau})$ and $\ell_f(\boldsymbol{x})$, are smooth functions which encode the reorientation tasks. To reorient the body, the running/terminal cost can be chosen as,
    \begin{equation}
        \begin{aligned}
            \ell(\boldsymbol{x}, \boldsymbol{\tau})&=e(\boldsymbol{\Theta}_d,\boldsymbol{\Theta})+\frac{1}{2}\boldsymbol{u}_f^T\boldsymbol{Q}_{u_{f}}\boldsymbol{u}_f + \frac{1}{2}\boldsymbol{\tau}^T\boldsymbol{R}_{\tau}\boldsymbol{\tau}\\
            \ell_f(\boldsymbol{x})&=w \cdot e(\boldsymbol{\Theta}_d,\boldsymbol{\Theta}),
        \end{aligned}
    \end{equation}
    where the attitude error $e(\cdot,\cdot)$ function (as used in \cite{Taeyoung2010}) between the current body orientation and the desired orientation $\boldsymbol{\Theta}_d$ is defined as
    \begin{equation}
        e(\boldsymbol{\Theta}_d,\boldsymbol{\Theta}) = \frac{1}{2}\text{tr}(\boldsymbol{I}-\boldsymbol{R}^T(\boldsymbol{\Theta}_d)\boldsymbol{R}(\boldsymbol{\Theta})),
    \end{equation}
    where $\boldsymbol{R}(\boldsymbol{\Theta})$ is the rotation matrix corresponding to the quaternion and $w$ is the weight for the final cost of the orientation ($w=500$ in this paper). $\boldsymbol{Q}_{u_{f}}$ and $\boldsymbol{R}_{\tau}$ are positive semi-definite matrices for the regularization on the velocities and tail torques, respectively. As the translation of body CoM is not of interest during reorientation, the diagonal elements of $\boldsymbol{Q}_{u_{f}}$ corresponding to $\boldsymbol{\dot{p}}$ are set as zeros.
    \subsubsection{System Dynamics Constraint}
    The dynamical feasibility is enforced by the forward Runge-Kutta (\textit{RK4}) integration of the system dynamics in the rollout of the DDP method.
    \subsubsection{Tail Joint Limitations}
    In the tailed quadrupedal system, the workspace of the tail is limited within a cone zone in Cartesian space to avoid the self-collision with the body and legs. Hence the joint limitations of tail must be considered to avoid self-collisions,
    \begin{equation}
        \boldsymbol{f}(\boldsymbol{q}_t) \in \mathcal{X}_t,
    \end{equation}
    where $\boldsymbol{f}$ is the forward kinematics of the tail and $\mathcal{X}_t$ is the feasible set of tail positions.
    \subsubsection{Tail Actuation Limitations}
    The motor torques of the tail are also limited, which are piece-wise box constraints in the inputs
    \begin{equation}
        \tau_{min} \leq \boldsymbol{\tau}_k \leq \tau_{max}.
    \end{equation}
DDP is able to efficiently solve trajectory optimization problem through the parameterized control trajectory. The constraints in the problem are handled with Augmented Lagrangian approach and relaxed barrier function sequentially in a hybrid framework \cite{HM-DDP}. The linear feedback policy along the optimal solution returned by the DDP solver can also be used to stabilize the trajectory tracking in the flight phase.
\subsection{Flight Controller}
In the flight phase, the optimized trajectory will be tracked with a time-varying linear feedback controller. The feedback tracking controller is in form of
\begin{equation}
    \boldsymbol{\tau} = \boldsymbol{\tau}_{f}^{ref} + \boldsymbol{K}_{p}(\boldsymbol{q}_f-\boldsymbol{q}_{f}^{ref})+\boldsymbol{K}_{d}(\boldsymbol{u}_f-\boldsymbol{u}_{f}^{ref}),
\end{equation}
where $\boldsymbol{\tau}_{f}^{ref}$, $\boldsymbol{q}_{f}^{ref}$ and $\boldsymbol{u}_{f}^{ref}$ are the optimized reference torques and reference joint trajectories obtained in the offline TO stage. $\boldsymbol{K}_p$ and $\boldsymbol{K}_d$ are proportional and derivative gains obtained from the feedback terms returned by DDP approach as mentioned in previous subsection. In addition, joint PD controllers are used to maintain leg configuration in airborne. 


After the body orientation is adjusted to the neighborhood of the desired orientation or the body descends to a certain height, the tail will be retracted to its minimum length quickly. Within the time duration of tail retraction, the robot legs are controlled by joint position controllers for landing preparation. Compared with the extended tail, the tail retractability makes the system CoM stay close to the geometric center of the support polygon, which alleviates the uneven force distribution of the feet in contact. Hence, the telescoping DoF turns to be important for the practical usage of appendages in falling quadruped robots.

\subsection{Stance Controller}
Once contact is detected, the system will switch to the stance controller. When the body orientation is well adjusted near the horizontal, less effort is needed to design a stance control strategy. To verify the feasibility of the proposed control framework, we employ a simple compliant joint PD controller to maintain each leg's configuration and keep the system balance in the stance phase. More advanced stance control (e.g., \cite{SHJ}) will be our interest in the future work.

\section{Simulation Validation}
We first evaluated the proposed system integration and control framework in MuJoCo \cite{mujoco}, which is a high-fidelity physic engine. 
The simulator run at $1000$ Hz, where the system forward dynamics was simulated and the contact between the foot and ground was detected. 
The friction coefficient was set as $\mu=0.8$. 
In simulation, the tailed A1 robot was simulated to reorient its body angle and land safely from various initial orientations with a falling height of $1.85$ m ($\boldsymbol{p}_z = 1.85$ m).
We assumed that the tailed A1 robot started from a static state in all simulations.

\subsection{Aerial Reorientation}
\begin{figure}[tb]
    \centering
    \includegraphics[width=80mm]{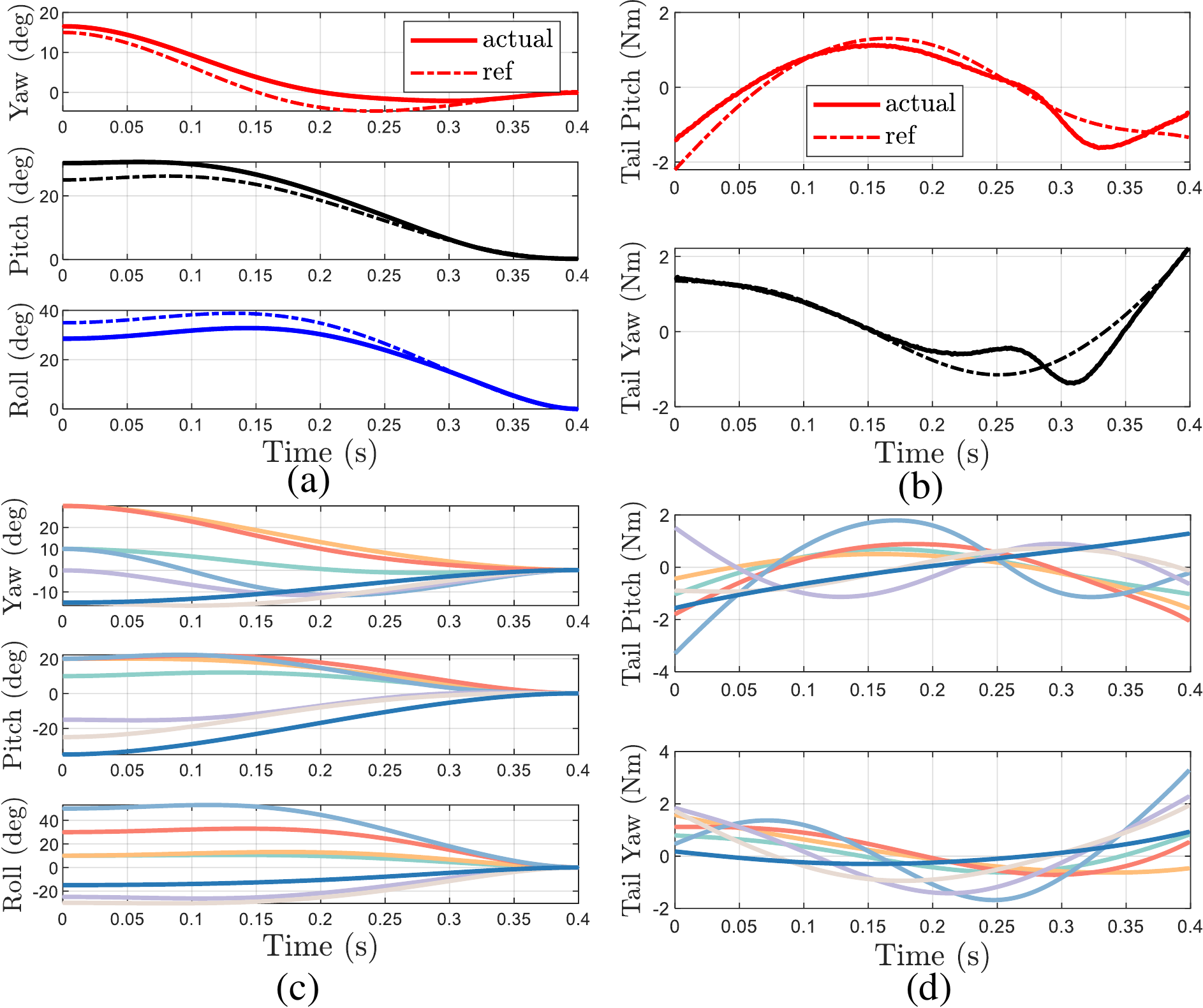}
    \caption{(a-b) Simulation results of aerial reorientation in the flight phase within $0.4$ s and the initial orientation is $[15^{\degree},25^{\degree},35^{\degree}]$. (c-d) Body orientation and joint torques in a bunch of simulations with different initial orientations and curves in the same color correspond to the same simulation.}
    \label{fig:aerial_reorientation}
\end{figure}
In the offline trajectory optimization stage, the desired orientation $\boldsymbol{\Theta}_d$ was set as $[0,0,0,1]$ and the time budget for the reorientation task was $0.4$ s with $N=200$. The forward dynamics of tSRBD in TO was implemented using the \textit{spatial-v2} package \cite{spatialv2} in {MATLAB}. The nonlinear optimization problem was then solved with a custom-made DDP solver \cite{HM-DDP}, where \textit{casadi} \cite{Andersson2019} was used as a tool of auto-differentiation for the computations of derivatives of the forward dynamics, objective functions, and constraints. Solving such an optimization problem with zero controls as initial guess usually took between 100 and 200 iterations. 
The optimized results ($\boldsymbol{\tau}_f^{ref}$, $\boldsymbol{q}_f^{ref}$ and $\boldsymbol{u}_f^{ref}$) were then interpolated with polynomials as reference inputs for the tracking controller. 

To verify the aerial reorientation capability, the tailed A1 robot fell with various initial body orientations in simulator. To give an intuitive visualization, the orientation was plotted in Euler angles (in \textit{yaw-pitch-roll}). The simulation results with an initial orientation of $[15^{\degree},25^{\degree},35^{\degree}]$ were shown in Fig.~\ref{fig:aerial_reorientation}(a-b). The optimized trajectory (dashed line) was well tracked and the body attitude was adjusted to the desired one, even though model errors (e.g., $1.4\times$ tail mass) were introduced manually. Simulation results with other different initial body orientations were presented in Fig.~\ref{fig:aerial_reorientation}(c-d). These results demonstrated the robot's 3D reorientation capability.

\subsection{Consecutive Motion}
\begin{figure}[t]
    \centering
    \includegraphics[width=85mm]{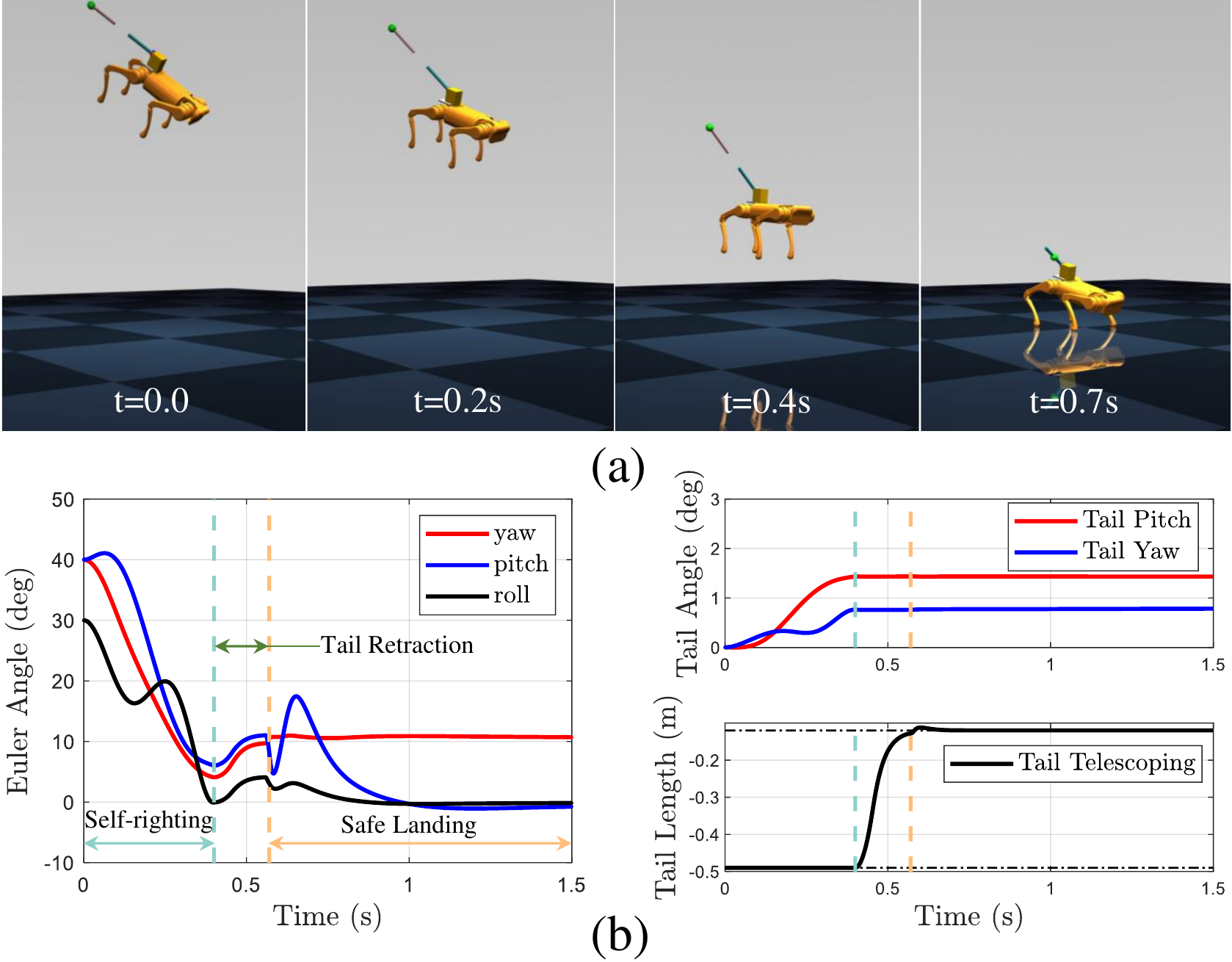}
    \caption{(a) Snapshots of the consecutive motion in MuJoCo environment. (b) Body orientation and tail motion over time.}
    \label{fig:full_res}
\end{figure}

To validate the consecutive motions of aerial orientation and safe landing, one trial was shown in Fig. \ref{fig:full_res}. The robot was dropped with an initial Euler angle of $[40^{\degree},40^{\degree},30^{\degree}]$. 
The tailed robot adjusted its body orientation by swinging the tail within the first $0.4$ s then retracted the tail for landing. The contact was detected at $0.56$ s and the system switched to the stance controller for keeping balance. From Fig.~\ref{fig:full_res}(b), there were body orientation errors at touch-down because of the disturbance caused by tail retraction. Small errors in pitch and roll (several degrees) can be eliminated by the landing control after the robot was settled down. To eliminate the error in yaw, three DoFs of the tail can be activated together under proper control, which will be our future study.



\begin{table}[b]
	\centering
	\caption{Tailed A1 Robot and Test Platform Parameters}\label{table:a1param}
	\begin{tabular}{p{2.4cm} p{0.8cm} p{3.5cm}}
		\toprule  
		\textbf{Parameters} &\textbf{Symbol} &\textbf{Value} \\ 
		\midrule
		A1 Robot Mass      &$m_b$     &$12.45$ kg\\
		\midrule
		A1 Robot Inertia   &$\boldsymbol{I}_b$ &$[0.12, 0.39, 0.45]$ kg$\cdot$m$^2$\\
		\midrule
		Test Body Mass   &$m^t_{b}$  &$11.5$ kg \\
		\midrule
		Test Body Inertia    &$\boldsymbol{I}^t_b$ &$[0.05,0.25,0.22]$ kg$\cdot$m$^2$\\
		\midrule 
		Tail Mass      &$m_t$     &$1.25$ kg\\
		\midrule
		Tail Length Range     &$\ell_t$  &$[0.12,\;0.49]$ m\\
		\bottomrule 
	\end{tabular}
\end{table}

\section{Experimental Results}
\subsection{Experimental Setup}
A $2.3$ kg 3-DoF robotic tail prototype ($850$ g tail base package, $820$ g tail scissor linkages, and $630$ g tail mass end) was integrated into the Uniree A1 robot.
The tail base was placed above the middle of the robot's two hinder leg hip actuators.
The tail can provide a large range of motion (${- 90^{\degree}\sim180^{\degree}}$ in pitch, $\pm 180^{\degree}$ in yaw, and $0.12$ $\sim$ $0.49$ m in length).
The tail end mass includes a T-motor Antigravity 5008 KV170 ($128$ g Incl. Cable, open-source hardware VESC as the motor driver) and a worm gearbox (gear ratio $10:1$), which were in charge of controlling the tail length.
The tail's pitch/yaw motion was controlled by the tail base that consists of two T-motor AK60-6 ($315$ g) and a differential bevel gear gearbox.
The differential actuation mechanism can provide a large range of motion and large output torque.
Other electrical components (a $400$ g battery, a Raspberry Pi 3B+ control board, and an LPMS-BE1 IMU module) were placed on the bottom of Unitree A1.
\begin{figure}[bt]
    \centering
    \includegraphics[width=70mm]{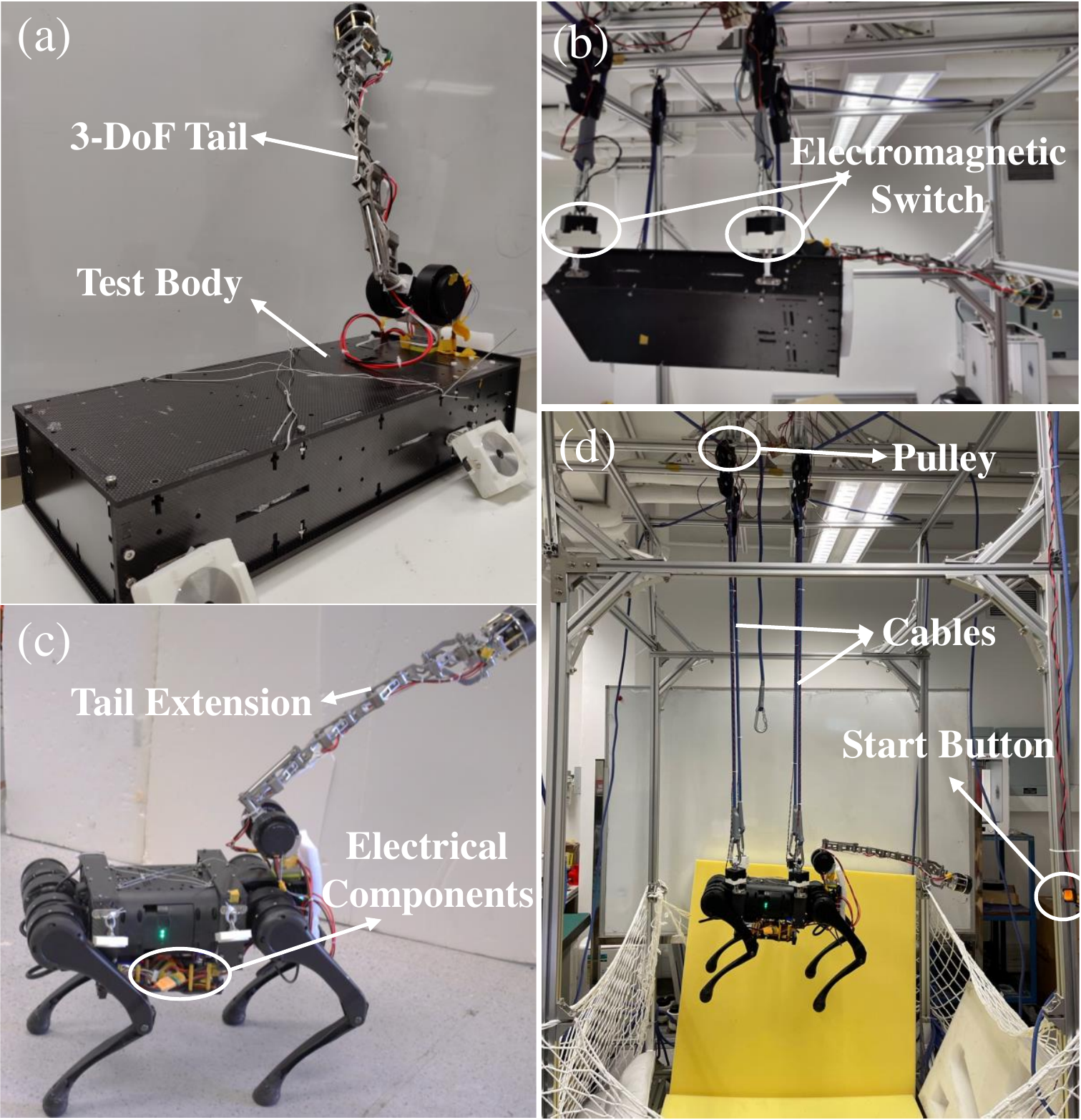}
    \caption{Experimental platforms including a flight-phase test platform and a tailed A1 robot. (a) Flight-phase test platform. (b) The test platform is suspended for drop tests. (c) A1 robot with an extended tail. (d) The tailed A1 robot is suspended for drop tests.}
    \label{fig:experimental_setup}
\end{figure}

To verify the tailed quadruped robot's aerial reorientation and landing capability safely and repeatably, we built up an auxiliary truss-structured platform for hanging and releasing the robot as shown in Fig.~\ref{fig:experimental_setup}(d). The tailed robotic system was suspended by four cables via electromagnetic holders. 
By adjusting the length of each cable, the initial body orientation and height can be set as desired. Once the start button (Fig.~\ref{fig:experimental_setup}(d)) was pressed, the electromagnets de-energized and the controller was activated at the same time.

Since repeated dropping experiments may damage the motors of the quadrupedal robot, we designed a flight-phase test platform (Fig.~\ref{fig:experimental_setup}(a)) to test the aerial reorientation function initially. The flight-phase test platform consisted of a cuboid body and the same 3-DoF tail. The physical parameters of the test platform were given in Table \ref{table:a1param}. We mainly repeated the aerial reorientation experiments on the test platform and then transferred to the tailed A1 robot with fine tuning. The body orientation and angular velocity were estimated from the internal IMU. A contact was detected by a sudden acceleration change in the vertical direction. A soft cushion was also laid on the ground to protect the robots.

\subsection{Flight-Phase Test Platform Experiments}
\begin{figure}[tb]
    \centering
    \includegraphics[width=85mm]{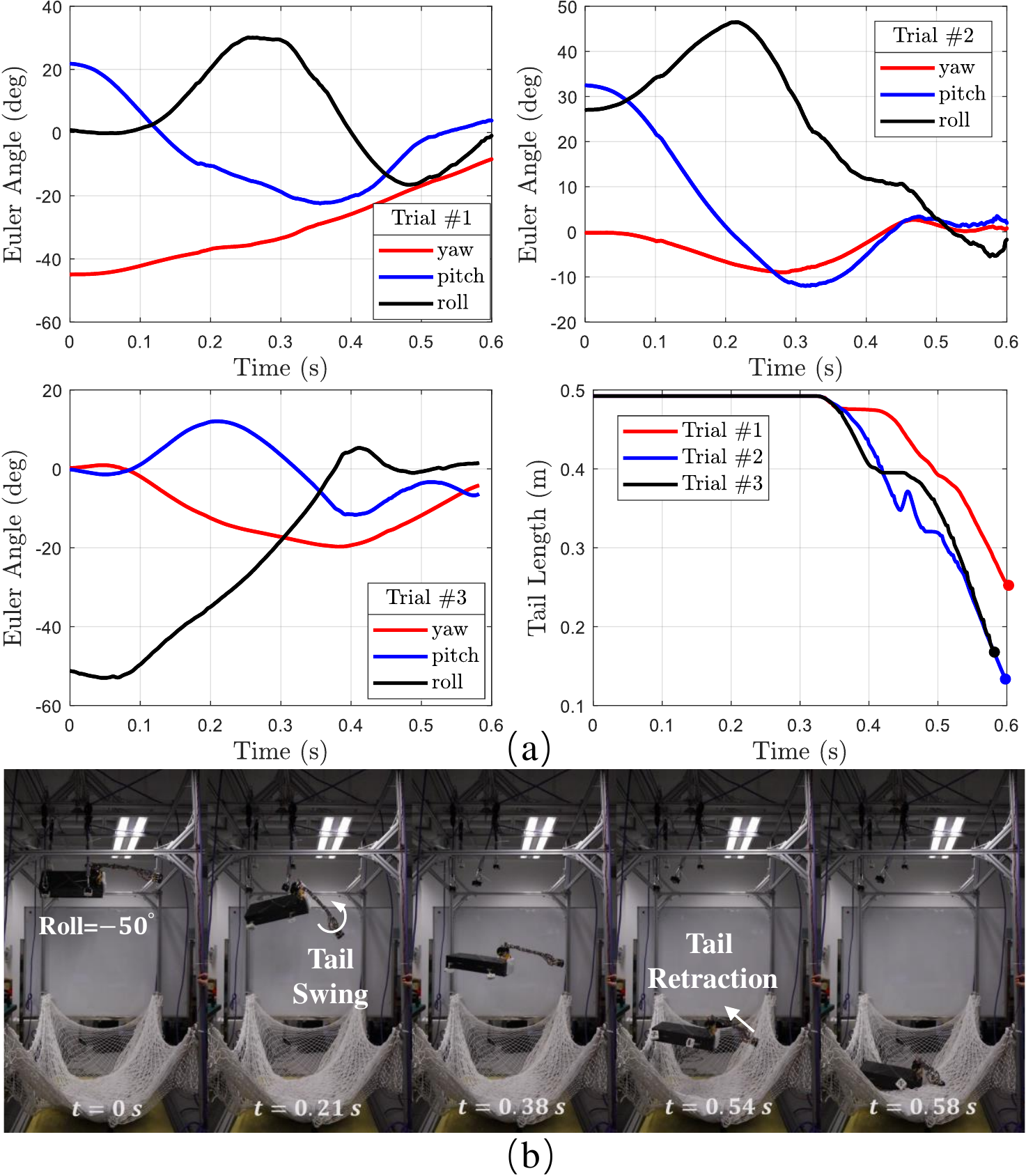}
    \caption{(a) Experimental results of three trials on the flight-phase test platform: Euler angle of body orientation and tail length variations versus time. (b) Motion snapshots of trial $\#3$.}
    \label{fig:aerial_exp}
\end{figure}

To validate the tail can increase the quadrupedal robot's 3D aerial righting capability for safe landing, we dropped the test platform from various initial orientations onto the cushion from $1.85$ m height. 
The initial body orientation was manually adjusted and the tail was kept in its zero joint configuration with maximal length. 
With the initial orientation, an optimized trajectory and tracking controller were offline planned as in discussed in Section III. To handle the model uncertainties, an additional feedback PD controller is hand-tuned to improve the tracking performance. We show the experimental results of three trials in Fig.~\ref{fig:aerial_exp}(a). The platform fell from three totally different initial orientations and the final orientations were successfully adjusted to the neighbourhood of the desired orientation at the end of the flight phase.
The observed errors are tolerable (within $\pm10^{\degree}$) for quadrupedal robots' landing. Especially, in the third trial, the initial roll offset was up to $-50^{\degree}$, which was a challenging orientation for common quadrupedal robots to recover from in falling. The motion snapshots of trial 3 was shown in Fig.~\ref{fig:aerial_exp}(b).
In the experimental results, we can see the tail started to retract its length at $0.35$ s in airborne and kept retracting till touch down. The tail retraction speeds in experiments were repeatable ($\sim 2$ m/s, estimated from Fig.~\ref{fig:aerial_exp}(a)) . 
In the experiments, the tail retraction sometimes got stuck because the tail's fast swing speed created large centrifugal force and the tail telescoping motor attained its torque limits.

\subsection{Tailed A1 Robot Experiments}
\begin{figure}[tb]
    \centering
    \includegraphics[width=85mm]{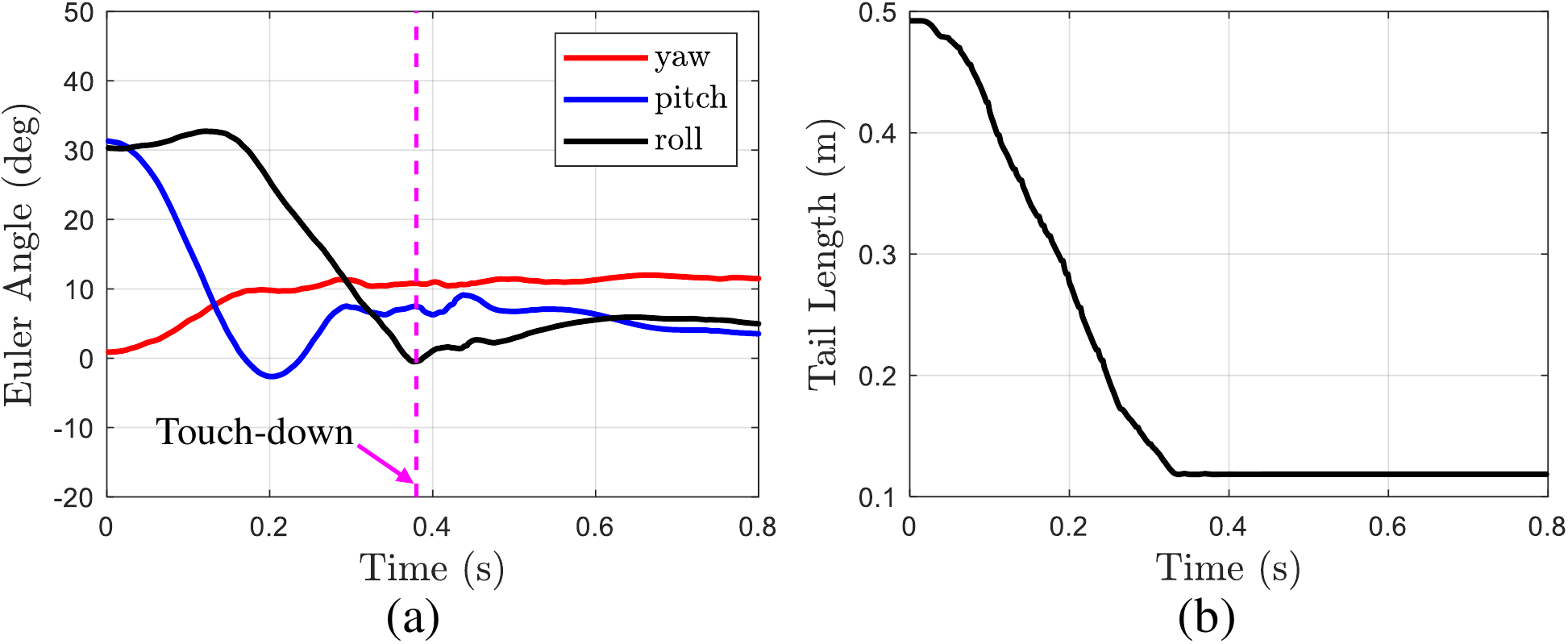}
    \caption{Experimental results on the tailed A1 robot, including body Euler angle and tail length change. A snapshot is shown in Fig. 1.}
    \label{fig:full_exp}
\end{figure}

To validate a consecutive large 3D reorientation and safe landing on a tailed quadruped robot, the tailed A1 robot was dropped from a non-negligible initial body angle (Fig.~1). The initial body angle was $[0^{\degree},30^{\degree},30^{\degree}]$ and the desired orientation was $[0^{\degree},0^{\degree},0^{\degree}]$. A relatively low height, 1 m, was selected for dropping, because the robot would not suffer from too large motor current. This safety-oriented height selection did not affect our goal of demonstrating feasibility. Limited by the flight duration, the tail control strategy was slightly different and the tail would retract earlier for a trade-off between the functions of tail swing and retraction. At $0.38$ s, the robot touched the ground and the tail retracted to its minimum length (Fig. \ref{fig:full_exp}). The body orientation almost converged to the horizontal plane, although the yaw angle was around $10^{\degree}$. One of reasons is that the tail retraction was not considered during reorienting the body. As shown in Fig. 1, the robot can land stably even with the small orientation error. The same trial without retracting the tail was conducted and a robot falling was observed (see video). This further emphases the importance of the telescoping DoF.

\section{Discussion}
We have successfully used the 3-DoF tail for quadruped robot's 3D aerial reorientation and further safe landing. However, the tail usage in this paper is still straightforward. To show the proof of concept, the 3-DoF tail degenerated to a 2-DoF one before body self-righting and the telescoping function was only used for landing preparation. Actually, the telescoping DoF can be involved in the whole flight phase, which may generate more effective reorientation trajectories towards more robust and safe landing. Although this practice requires a new model, it can eliminate the disturbance of tail retraction that occurs currently.  In the landing control, we also used a simple PD controller since small orientation offset was achieved. More advanced landing planning/control like contact-aware trajectory optimization can be introduced and it can fully make use of legs' control authority to improve landing success. Lastly, we admit that the introduction of the tail would increase the total mass and may affect robot walking, but we did not directly change the foot design as \cite{KurtzMiniCheetah2022}. The tail package weight can be further reduced by optimizing the tail scissor linkage structures.





\section{Conclusion and Future Work}
In this paper, we proposed to integrate a 3-DoF tail module into a falling quadruped robot, enabling 3D aerial reorientation capability for safe landing.
The simplified robot dynamic model was presented and we also proposed a simple but effective control framework to demonstrate the feasibility of the system integration. 
A flight-phase test platform with comparable inertial properties to the quadrupedal robot (Unitree A1) was built for initial experimental verification, demonstrating the tail's effectiveness on 3D body reorientation and fast retractability during falling.
A consecutive large 3D reorientation and safe landing motion were successfully completed on the tailed A1 robot. In the future, besides addressing the concerns mentioned in Discussion, we plan to investigate the advantages of the 3-DoF tail on assisting the quadrupedal locomotion, such as accelerating/decelerating and turning sharply. Besides, the tail's telescoping function can be used for simple interactions with the environment, facilitating the deployment in real world.


\begin{thebibliography}{}
\bibitem{KaneFallingCatPhenomenon1969}
T. R. Kane and M. P. Scher. ``A dynamical explanation of the falling cat phenomenon," {\em International Journal of Solids and Structures}, Vol. 5, No. 7, pp. 667-670, 1969.

\bibitem{FukushimaSquirrels2021}
Toshihiko Fukushima, Robert Siddall, Fabian Schwab, Severine L. D. Toussaint, Greg Byrnes, John A. Nyakatura and Ardian Jusufi. ``Inertial Tail Effects during Righting of Squirrels in Unexpected Falls: From Behavior to Robotics," {\em Integrative and Comparative Biology}, Vol. 61, No. 2, pp. 589-602, 2021.

\bibitem{SHJ}
S. H. Jeon, S. Kim and D. Kim, "Online Optimal Landing Control of the MIT Mini Cheetah," 2022 International Conference on Robotics and Automation (ICRA), pp. 178-184, 2022.


\bibitem{KurtzMiniCheetah2022}
TVince Kurtz, He Li, Patrick M. Wensing, and Hai Lin. ``Mini Cheetah, the Falling Cat: A Case Study in Machine Learning and Trajectory Optimization for Robot Acrobatics," {\em Proceeding of IEEE International Conference on Robotics and Automation}, pp. 4635-4641, 2022.

\bibitem{RudinCatLike2022}
Nikita Rudin, Hendrik Kolvenbach, Vassilios Tsounis, and Marco Hutter. ``Cat-Like Jumping and Landing of Legged Robots in Low Gravity Using Deep Reinforcement Learning," {\em IEEE Transactions on Robotics}, Vol. 38, No. 1, pp. 317-328, 2022.



\bibitem{GosselinReorientation2022}
Mark Charlet and Clément Gosselin. ``Reorientation of Free-Falling Legged Robots ," {\em ASME Open J. Engineering}, No. 1, 011009, 2022.

\bibitem{KolvenbachMoon2019}
Hendrik Kolvenbach, Elias Hampp, Patrick Barton, Radek Zenkl and Marco Hutter. ``Towards Jumping Locomotion for Quadruped Robots on the Moon," {\em Proceeding of IEEE/RSJ International Conference on Intelligent Robots and Systems}, pp. 5459-5466, 2019.

\bibitem{ZacharyCMU2022}
Chi-Yen Lee, Shuo Yang, Benjamin Bokser, and Zachary Manchester. ``Reaction Wheel Assisted Locomotion for Legged Robots," {\em ICRA Workshop on Legged Robots}, 2022.

\bibitem{Roscia2022}
Francesco Roscia, Andrea Cumerlotti, Andrea Del Prete, Claudio Semini, and Michele Focchi. ``Orientation Control System: Enhancing Aerial Maneuvers
for Quadruped Robots," {\em https://arxiv.org/abs/2208.12066}, 2022.

\bibitem{Liu2021}
Liu, Y, Ben-Tzvi, P, ``Feedback Control of the Locomotion of a Tailed Quadruped Robot." {\em ASME International Design Engineering Technical Conferences and Computers and Information in Engineering Conference}, 2021.

\bibitem{YangCMU2021}
Yanhao Yang, Joseph Norby, Justin K. Yim, and Aaron M. Johnson. ``Improving Tail Compatibility Through Sequential Distributed Model Predictive Control," {\em RSS Workshop on Software Tools for Real-Time Optimal Control}, 2021.

\bibitem{FawcettArticulatedTails2021}
R. T. Fawcett, A. Pandala, J. Kim, and K. Akbari Hamed. ``Real-Time Planning and Nonlinear Control for Quadrupedal Locomotion With Articulated Tails," {\em Journal of Dynamic Systems, Measurement}, Vol. 143, No. 7, 2021.

\bibitem{BriggsTails2012}
Randall Briggs, Jongwoo Lee, Matt Haberland, and Sangbae Kim. "Tails in biomimetic design: Analysis, simulation, and experiment," {\em IEEE/RSJ International Conference on Intelligent Robots and Systems}, pp. 1473-1480, 2012.

\bibitem{YangCMU2022}
Yanhao Yang, Joseph Norby, Justin K. Yim, and Aaron M. Johnson. ``Proprioception and Tail Control Enable Extreme Terrain Traversal by Quadruped Robots," {\em ICRA Workshop on Legged Robots}, 2022.

\bibitem{NorbyAerodynamic2021}
Joseph Norby, Jun Yang Li, Cameron Selby, Amir Patel, and
Aaron M. Johnson. ``Enabling Dynamic Behaviors With Aerodynamic Drag in Lightweight Tails," {\em IEEE Transactions on Robotics}, Vol. 37, No. 4, pp. 1144-1153, 2021.



\bibitem{LiuSerpentine2021}
Yujiong Liu and Pinhas Ben-Tzvi. ``Dynamic Modeling, Analysis, and Design Synthesis of a Reduced Complexity Quadruped with a Serpentine Robotic Tail," {\em Integrative and Comparative Biology}, Vol. 61, No. 2, pp. 464-477, 2021.

\bibitem{LiuICRA2022}
Yujiong Liu and Pinhas Ben-Tzvi. ``Systematic Development of a Novel, Dynamic, Reduced Complexity Quadruped Robot Platform for Robotic Tail Research," {\em Proceeding of IEEE International Conference on Robotics and Automation}, pp. 4664-4670, 2022.

\bibitem{Heim2016}
Steve W. Heim, Mostafa Ajallooeian, Peter Eckert, Massimo Vespignani, and Auke Jan Ijspeert. ``On designing an active tail for legged robots: simplifying control via decoupling of control objectives," {\em Industrial Robot: An International Journal}, Vol. 43, No. 3, pp. 338-346, 2016.

\bibitem{Patelturning2013}
Amir Patel and M. Braae. ``Rapid turning at high-speed: Inspirations from the cheetah's tail," {\em Proceeding of IEEE/RSJ International Conference on Intelligent Robots and Systems}, pp. 5506-5511, 2013.


\bibitem{ChuNSA2019}
Xiangyu Chu, Chun Ho David Lo, Carlos Ma, and Kwok Wai Samuel Au. ``Null-Space-Avoidance-Based Orientation Control Framework for Underactuated, Tail-Inspired Robotic Systems in Flight Phase,'' \textit{IEEE Robotics and Automation Letter}, Vol. 4, No. 4, pp. 3916-3923, 2019.


\bibitem{ChuNSA2022}
Xiangyu Chu, Chun Ho Lo, Tommaso Proietti, Conor J. Walsh, and Kwok Wai Samuel Au. ``Opposite Treatment on Null Space: A Unified Control Framework for a Class of Underactuated Robotic Systems With Null Space Avoidance,'' \textit{IEEE Transactions on Control Systems Technology}, early access, 2022.

\bibitem{Taeyoung2010}
Taeyoung Lee, Melvin Leok, and N Harris McClamroch. ``Geometric tracking control of a quadrotor uav on se(3)". In IEEE Conference on Decision and Control, pp. 5420–5425, 2010.

\bibitem{AnMorphable2020}
Jiajun An, TY Chung, Chun Ho David Lo, Carlos Ma, Xiangyu Chu, KW Samuel Au. ``Development of a Bipedal Hopping Robot With Morphable Inertial Tail for Agile Locomotion,'' \textit{ IEEE RAS/EMBS International Conference for Biomedical Robotics and Biomechatronics (BioRob)}, pp. 132-139, 2020.

\bibitem{AnMorphable2022}
Jiajun An, Xin Ma, Chun Ho David Lo, Weeshen Ng, Xiangyu Chu, and Kwok Wai Samuel Au. ``Design and Experimental Validation of a Monopod Robot With 3-DoF Morphable Inertial Tail for Somersault,'' \textit{IEEE/ASME Transactions on Mechatronics}, early access, 2022.

\bibitem{spatialv2}
[Online]. Available: http://royfeatherstone.org/spatial/v2/

\bibitem{RDLN}
[Online]. Available: \text{https://ethz.ch/content/dam/ethz/special-interest/}
\text{mavt/robotics-n-intelligent-systems/rsl-dam/documents/RobotDynamics}
\text{2017/RD\_HS2017script.pdf.}  \textit{Robot Dynamics Lecture Notes}, Robotic Systems Lab, ETH Zurich.

\bibitem{HM-DDP}
Yunix Tang, Xiangyu Chu, K. W. Au, ``HM-DDP: A Hybrid Multiple-shooting Differential Dynamic Programming Method for Constrained Trajectory Optimization", arXiv:2109.07131, 2021.

\bibitem{Andersson2019}
Andersson, Joel AE, et al. ``CasADi: a software framework for nonlinear optimization and optimal control." {\em Mathematical Programming Computation}, 11.1, 1-36, 2019.

\bibitem{mujoco}
E. Todorov, T. Erez and Y. Tassa, ``MuJoCo: A physics engine for model-based control," 2012 IEEE/RSJ International Conference on Intelligent Robots and Systems, 2012, pp. 5026-5033, doi: 10.1109/IROS.2012.6386109.

\bibitem{A1}
[Online]. Available: https://shop.unitree.com/products/unitree-a1.
\end{thebibliography}
\end{document}